\pdfoutput=1

\documentclass[11pt]{article}

\usepackage[]{acl}

\usepackage{times}
\usepackage{latexsym}
\usepackage{booktabs, multirow} 

\usepackage[T1]{fontenc}

\usepackage[utf8]{inputenc}

\usepackage{microtype}

\usepackage{inconsolata}
\usepackage{graphicx}

\usepackage{enumitem}
\usepackage[edges]{forest}

\title{A Case Study on Filtering for End-to-End Speech Translation}

\author{Md Mahfuz Ibn Alam$^\alpha$ \qquad Antonios Anastasopoulos$^{\alpha,\gamma}$ \\
        $^\alpha$Department of Computer Science, George Mason University \hfill \\
        $^\gamma$Archimedes AI Research Unit, RC Athena, Greece\\
        \texttt{\{malam21,antonis\}@gmu.edu}}

\begin{document}
\maketitle
\begin{abstract}
It is relatively easy to mine a large parallel corpus for any machine learning task, such as speech-to-text or speech-to-speech translation. Although these mined corpora are large in volume, their quality is questionable. This work shows that the simplest filtering technique can trim down these big, noisy datasets to a more manageable, clean dataset. We also show that using this clean dataset can improve the model's performance, as in the case of the multilingual-to-English Speech Translation (ST) model, where, on average, we obtain a 4.65 BLEU score improvement.
\end{abstract}

\section{Introduction}
End-to-end speech translation (E2E ST) represents a paradigm shift from cascaded speech translation by utilizing a direct model for converting source language speech into target-language text. This approach offers advantages such as low latency and avoiding the "error propagation" issue often encountered in traditional cascade methods~\cite{DBLP:journals/corr/WeissCJWC17, sperber-paulik-2020-speech}.  In recent times, end-to-end speech translation (ST) models have demonstrated noteworthy progress, achieving comparable or, in some instances, superior results when compared to traditional cascaded ST models~\cite{bentivogli-etal-2021-cascade, anastasopoulos-etal-2021-findings, anastasopoulos-etal-2022-findings}. 

Despite these benefits, E2E ST models face challenges when compared to the cascaded methods~\cite{di-gangi-etal-2019-must}, as there are robust training data available for automatic speech recognition (ASR) and machine translation (MT) which can be employed in cascade methods. The relatively limited training data for E2E ST models may lead to sub-optimal performance in certain scenarios. Prior research endeavors frequently tap into machine translation (MT) data to enhance training by applying multi-task learning techniques, as observed in works by~\citet{ye-etal-2022-cross, tang-etal-2021-improving}. Here the encoder and decoder are shared between speech translation (ST) and machine translation (MT) and the model acquires similar representations from distinct modalities.

Another research direction can be automatically creating large-scale mined corpus~\cite{duquenne-etal-2023-speechmatrix}. However, the created data is often too noisy mostly because it is too hard to align data in two different languages. On top of that, for Speech Translation we have to also align between two different modalities. Then the research focus falls on filtering the noisy data. This work focuses on filtering Speech Translation data and observing if by doing so the systems' performance can be improved.

Our main contributions in a nutshell:
\begin{itemize}[noitemsep,nolistsep]
    \item We try different filtering methods for Speech Translation.
    \item We show that the simplest ratio-based filtering can effectively differentiate clean data from noisy data.
    \item We show that a somewhat good ST model trained by a small clean data can efficiently filter a noisy corpus and thus improve the models' performance.
\end{itemize}


\begin{figure}[!t]\centering
\includegraphics[scale=0.8]{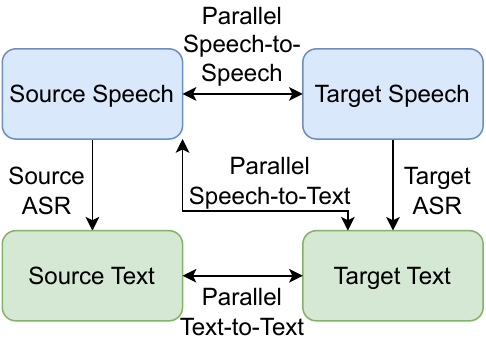}
\caption{Pipeline of our noisy data creation.}
\label{fig:Approach}
\end{figure}

\section{Approach}
Our approach is of three parts: Firstly, we create noisy speech-text data pairs. Secondly, we score each pair using our sorting techniques. Thirdly, we filter the pairs based on the scores and create subsets of pairs to train the model.
\subsection{Noisy Data Creation}
\begin{table}[!t]
    \centering
    \scriptsize
    \begin{tabular}{l|ccc}
        \toprule
        \textbf{Language} & \textbf{Model} & \textbf{WER $\downarrow$}& \textbf{Dataset}\\
        \midrule
        \multirow{3}{*}{English (EN)} & ~\cite{di-gangi-etal-2019-must} & 25.81 & \multirow{3}{*}{MuST-C} \\
        & Monolingual & 14.26 & \\
        & Multilingual & - & \\
        \midrule
        \multirow{2}{*}{French (FR)} & ~\cite{DBLP:journals/corr/abs-2102-01757} & 45.6 & \multirow{2}{*}{MTedX} \\
        & Monolingual & 19.2 & \\
        & Multilingual & 22.24 & \\
        \midrule
        \multirow{2}{*}{Spanish (ES)} & ~\cite{DBLP:journals/corr/abs-2102-01757} & 46.4 & \multirow{2}{*}{MTedX} \\
        & Monolingual & 17.06 & \\
        & Multilingual & 18.17 & \\
        \midrule
        \multirow{2}{*}{Portuguese (PT)} & ~\cite{DBLP:journals/corr/abs-2102-01757} & 54.8 & \multirow{2}{*}{MTedX} \\
        & Monolingual & 22.86 & \\
        & Multilingual & 24.48 & \\
        \midrule
        \multirow{2}{*}{Italian (IT)} & ~\cite{DBLP:journals/corr/abs-2102-01757} & 48.0 & \multirow{2}{*}{MTedX} \\
        & Monolingual & - & \\
        & Multilingual & 21.37 & \\
        \bottomrule
    \end{tabular}
    \caption{WER score of the ASR systems used to transcribe the SpeechMatrix dataset. The multilingual model uses FR, ES, PT, and IT languages. The monolingual models outperform the multilingual models.}
    \label{tab:ASR}
\end{table}

For our work, we have used a speech-to-speech translation dataset, SpeechMatrix (SM). To create the speech-to-text dataset for ST training, we have used ASR models to transcribe both the source and target sides. Figure~\ref{fig:Approach} outlines our approach. We created an ASR system, using each language's MuST-C or MTedX dataset. The WER of these ASR systems is shown in Table~\ref{tab:ASR}. The speech-to-speech translation dataset itself is noisy, and this approach adds one more layer of noise as the ASR models are imperfect. We explore if filtering approaches can work on this super noisy data.

\subsection{Scoring Techniques}
For our experiments, we explore ratio-based scoring techniques and one NLL(Negative Log likelihood)-based technique. The ratio-based techniques are faster to compute than NLL-based ones.

\noindent \textbf{Text-Text Ratio} To calculate this score, we divide the number of tokens from the source side transcript sentence by the number of tokens from the target side transcript sentence.

\noindent \textbf{Speech-Text Ratio} This is the same as the Text-Text Ratio, but now, on the source side, instead of the token count of the transcript text, we use the length in seconds of the speech as the numerator.

\noindent \textbf{Speech-Speech Ratio} As our original data is Speech-to-Speech data, we can calculate this score where on both the source and target side, we find the length (audio seconds) to compute the ratio.

\noindent \textbf{Text-Speech Ratio} Here, on the source side, we calculate the number of tokens, and on the target side, we calculate seconds of speech.

\noindent \textbf{NLL(Negative Log Likelihood)-based} We have used our baseline systems to calculate this score. Given a speech-text data pair ${<}x,y{>}$, we use Equation~\ref{eq:1} to calculate the NLL loss. Here, $f_\theta$ is our baseline model.
\begin{equation}
    \mathcal{L}(\theta) = - \left( y \, log \, \hat{y}_\theta + (1 - y) \, log \, (1 - \hat{y}_\theta) \right) 
    \label{eq:1}
\end{equation}
\begin{equation}
    \hat{y}_\theta = \sigma(f_\theta(x))
    \label{eq:2}
\end{equation}
\subsection{Filtering}
After obtaining the score of each data instance using the techniques explained above, we use two approaches to select a subset of clean data. For ratio-based techniques, we calculate $z$-scores, and for NLL-based techniques, we use percentiles.

\noindent \textbf{$z$-score} calculation requires first determining the mean, $\mu$ and standard deviation, $\sigma$ of the ratio-based scores of all the instances. Once we have these, we can calculate the $z$-score using Equation~\ref{eq:3}. Here, $x$ is the ratio of a data instance. After calculating the $z$-score from each technique, we create four subsets where the $z$-score of the instances are less than or equal to 0.25, 0.5, 0.75, and 1.0. $z$-score here ranges from $0$ to $\infty$.
\begin{equation}
    z = |(x - \mu) / \sigma|
    \label{eq:3}
\end{equation}

\noindent \textbf{Percentile} calculation requires sorting the data instances based on NLL loss score. After sorting in ascending order, we create four subsets that take the first 20\%, 40\%, 60\%, and 80\% data instances.

\noindent \textbf{Union and Intersection}
We create another subset of data instances by performing a union and intersection operation between the best two subsets of each language pair.

\section{Experimental Settings}
\subsection{Datasets}
\noindent \textbf{MuST-C}~\cite{di-gangi-etal-2019-must} are multilingual speech translation corpora whose size and quality facilitate end-to-end ST training from English into German, Spanish, French, Italian, Dutch, Portuguese, Romanian and Russian. MuST-C includes at least 385 hours of audio recordings from TED Talks in English for each target language. They are automatically aligned with their manual transcriptions and translations on a sentence level.

\noindent \textbf{MTedX \cite{DBLP:journals/corr/abs-2102-01757}} supports speech recognition (ASR) and speech translation (ST) research across multiple non-English languages. The corpus is a collection of audio recordings from TEDx talks in 8 source languages. Transcriptions are segmented into sentences and aligned to the audio in the source language and the translation in the target language.

\noindent \textbf{SpeechMatrix \cite{duquenne-etal-2023-speechmatrix}} is a groundbreaking initiative that introduces a vast and extensive multilingual corpus focused on speech-to-speech translations (S2ST). This corpus consists of authentic speech recordings of the European Parliament, offering a rich and diverse collection. In total, SpeechMatrix boasts speech alignments across 136 language pairs, providing a comprehensive resource with a staggering cumulative duration of 418 thousand hours of speech data.

\subsection{Baseline Models}
We have created three baseline models. First, we have created a bilingual baseline model using the MuST-C or MTedX dataset. We consider these datasets as clean datasets. Thus, we expect these models' performance to be quite high in quality.  Next, we have created another baseline model using the SpeechMatrix dataset. This dataset is quite noisy; thus, we expect this model not to perform very well. Our last baseline model combines the MTedX or MuST-C datasets and the SpeechMatrix datasets without filtering. These baselines aim to quantify how adding the unfiltered SpeechMatrix affects the models' performance.

\subsection{ASR and ST Model}
For our experiments in ASR and ST, we use the same neural architecture. This setting allows us to initialize the ST encoder with the ASR encoder for faster convergence~\cite{9003774, bansal-etal-2019-pre}. We also created a multilingual ASR model and used it to initialize the multilingual ST model's encoder.  We used FAIRSEQ~\cite{ott-etal-2019-fairseq} to train end-to-end Transformer models for ST, using 80-dimensional log mel filterbank features with cmvn and SpecAugment~\cite{Park_2019}, and 2-d convolutions for downsampling of speech in time.  We use the \textit{s2t\_transformer\_s} architecture of FAIRSEQ, which has a network depth of 12 encoder layers and 6 decoder layers. All other hyper-parameters are discussed in the Appendix~\ref{sec:appendix_hyperparameter}.

\subsection{Language Pairs}
According to the clean data availability of the language pairs, we divided the language pairs that we experimented on into three categories: high-, mid-, and low-resource. High-resourced language pairs like English-French have a large amount of clean data as opposed to mid- or low-resourced. Thus, we can get a high-quality ST system in the high-resourced language pairs. In contrast, the models of the mid-resourced pairs like French-English and Spanish-Portuguese will have moderate quality. The low-resourced ones like Portuguese-Spanish will be even more degraded in quality.

\section{Results}
Using the above approach we create different subsets of data to train our systems with. We run experiments in two different settings: Bilingual, and Multilingual. This is to get an understanding of the filtering approaches' generalization ability.

\subsection{Bilingual Systems}
We have experimented on four language pairs.  Due to space constraints, all results of this experiment are in Appendix~\ref{sec:appendix_bilingual}. Table~\ref{tab:Summary} presents a summarized version of the results where we only show the highest-scoring models. We have two setups in each pair: trained only using the filtered SpeechMatrix data and trained by combining the filtered SpeechMatrix data with either MuST-C or MTedX.

\begin{table*}[!t]\centering
\footnotesize
\begin{tabular}{@{}l|l@{ }r@{ }r|l@{ }r@{ }r@{}}\toprule
\textbf{Lang Pair} &\textbf{ST System (Only SM)} &\textbf{\# of pairs} &\textbf{BLEU} &\textbf{ST System (MuST-C/MTedX + SM)} &\textbf{\# of pairs} &\textbf{BLEU}\\\midrule
\multirow{2}{*}{\textbf{En-Fr}} &Baseline (unfiltered SM) &1384112 &4.87 &Baseline (MuST-C) &269256 &31.51 \\
&80\% NLL &1107289 &\textbf{8.42} &MuST-C + SM (20\% NLL) &546078 &\textbf{32.07} \\\midrule
\multirow{2}{*}{\textbf{Fr-En}} &Baseline (unfiltered SM) &1403985 &5.20 &Baseline (MTedX + unfiltered SM) &1433931 &26.74 \\
&80\% NLL &1107289 &\textbf{6.31} &MTedX + SM (60\% NLL) &872337 &\textbf{27.66} \\\midrule
\multirow{3}{*}{\textbf{Es-Pt}} &Baseline (unfiltered SM) &1074027 &0.57 &Baseline (MTedX + unfiltered SM) &1094848 &26.28 \\
&\multirow{2}{*}{0.5 z Text-Text} &\multirow{2}{*}{559285} &\multirow{2}{*}{\textbf{5.13}} &\multirow{2}{*}[0.5em]{MTedX + } &\multirow{2}{*}{834039} &\multirow{2}{*}{\textbf{27.07}} \\
& & & &SM (0.5 z T-T $\bigcup$ 0.5 z Speech-Speech) & & \\\midrule
\multirow{2}{*}{\textbf{Pt-Es}} &Baseline (unfiltered SM) &1107283 &3.23 &Baseline (MTedX + unfiltered SM) &1118635 &22.64 \\
&60\% NLL &664369 &\textbf{4.38} &MTedX + SM (80\% NLL) &897178 &\textbf{23.7} \\
\bottomrule
\end{tabular}
\vspace{-.5em}
\caption{Results of filtering approaches for Bilingual ST systems. The models created using filtering approaches outperform the baseline in each language direction.}
\label{tab:Summary}
\vspace{-1em}
\end{table*}
\begin{table*}[!t]\centering
\footnotesize
\begin{tabular}{l|cccc|c|c}\toprule
\textbf{Multi ST System} &\textbf{ES-EN ($\Delta$)} &\textbf{FR-EN ($\Delta$)} &\textbf{PT-EN ($\Delta$)} &\textbf{IT-EN ($\Delta$)} &\textbf{Average ($\Delta$)} &\textbf{\# of pairs} \\\midrule
\textbf{~\cite{DBLP:journals/corr/abs-2102-01757}} &12.3 &12.0 &12.0 &10.7 &11.75 & 120538 \\\midrule
\textbf{(i) Baseline (MTedX)} &17.34 &18.92 &20.02 &15.74 &18.01 & 120538 \\
\textbf{(i) + SM (100\%)} &15.46 (-1.88) &19.93 (1.01) &16.61 (-3.41) &17.43 (1.69) &17.36 (-0.65) & 7292751\\
\textbf{(i) + SM (80\% NLL)} &17.04 (-0.30) &21.01 (2.09) &18.53 (-1.49) &18.59 (2.85) &18.79 (0.78) & 5858310 \\
\textbf{(i) + SM (60\% NLL)} &18.15 (0.81) &22.86 (3.94) &21.1 (1.08) &18.92 (3.24) &20.26 (2.25) & 4423867 \\
\textbf{(i) + SM (40\% NLL)} &19.86 (2.52) &24.76 (5.84) &22.35 (2.33) &20.51 (4.77) &21.87 (3.86) & 2989425 \\
\textbf{(i) + SM (20\% NLL)} &\textbf{20.4 (3.06)} &\textbf{24.96 (6.04)} &\textbf{24.45 (4.43)} &\textbf{20.81 (5.07)} &22.66 (4.65) & 1554982 \\
\bottomrule
\end{tabular}
\vspace{-0.5em}
\caption{Results of filtering approaches for Multilingual ST system. $\Delta$ is the difference from the baseline model.}
\label{tab:Multi}
\vspace{-2em}
\end{table*}

\noindent \textbf{High Resource (EN-FR)}
This language pair has the most clean data (270,000 instances) and the highest baseline score (31.51 BLEU) among the four pairs. This is the only pair where adding the whole noisy SpeechMatrix data substantially hurts the system's performance. All models created from five filtering techniques lead to higher BLEU scores than the Baseline (MuST-C + unfiltered SM). This shows that even a simple filtering technique is helpful instead of using all the noisy data.

We achieved the best performance (32.07 BLEU) by using only 20\% clean data from SpeechMatrix on top of the clean data, a 0.56 BLEU score improvement over the baseline model, Baseline(MuST-C). We find a similar trend for all filtering techniques, where adding more data hurts performance. Note that systems trained using just the SpeechMatrix are awful, yet our best model earns 3.55 BLEU points just by filtering compared to the Baseline(unfiltered SM) model.

\noindent \textbf{Mid Resource (FR-EN \& ES-PT)}
With data sizes of 30,000 and 21,000, respectively, FR-EN and ES-PT language pairs are mid-resourced. When trained using only the SpeechMatrix, the BLEU score of the best model for FR-EN increases by 1.11 BLEU, whereas ES-PT increases by 4.56 in comparison to the Baseline (unfiltered SM) system.

Combining the unfiltered SpeechMatrix data with the MTedX data for FR-EN gives us an additional 8-BLEU points boost compared to only using MTedX data. Filtering the SpeechMatrix with the NLL-based technique and using 60\% of the data again gives an additional boost of 0.92.

For ES-PT, the baseline score increases when combining the unfiltered SpeechMatrix data with the MTedX by 5.23 BLEU compared to Baseline (MTedX). Our best model is the union of our two best subsets (0.5 z Text-Text $\bigcup$ 0.5 z Speech-Speech), with an increase of 0.79~BLEU points over the Baseline (MTedX + unfiltered SM). Even using the intersection of our two best subsets, the system is better than the baseline one. This exhibits the prowess of the filtering techniques.

\noindent \textbf{Low Resource (PT-ES)}
This language pair has the lowest amount of clean data (11,000 instances). Thus, the baseline score (12.62 BLEU) is the lowest among the four pairs. Just adding the SpeechMatrix gives an astonishing 10.02 BLEU score improvement. Noisy parallel data apparently helps improve the system to a certain level when the models are low-resourced. Beyond that, showing the model noisy data only hurts the performance. Filtering is crucial in that aspect to trim down the noisy dataset to a cleaner version. The best model uses NLL-based filtering and 80\% of the dataset, which increases its performance by 1.06 BLEU points on top of the Baseline (MTedX + unfiltered SM).

\subsection{Multilingual Systems}
After observing the success of the NLL-based filtering for four language pairs, we have also explored a multilingual setup. Table~\ref{tab:Multi} shows the result of our multilingual experiments. The language directions are to English from Spanish, French, Portuguese, and Italian. Compared to the bilingual FR-EN results trained on the same dataset, we obtained a higher BLEU score (14.74 to 18.92).

We achieve even better scores by combining MTedX data with the first 20\% SpeechMatrix data filtered by NLL-loss for all language pairs. On average, this boosts our BLEU score by 4.65. We obtained the highest jump for FR-EN, a 6.04 BLEU score improvement. We obtain a similar trend for all other subsets, where the quality degrades the more noisy data we use.

\section{Conclusion}
All language pairs benefit from the simplest filtering techniques, from high-resourced to low-resourced pairs. The multilingual models' increase in model robustness is much more consistent and higher than that of bilingual ones.

\section{Limitations}
Even though we observed a huge increase in BLEU points by incorporating the simplest scoring and filtering techniques for all the language pairs that we have experimented with, one limitation arises when computing the NLL-based score. The limitation is that the computation time is huge. On the other hand ratio-based scoring techniques are faster by a huge margin. So, even if our best models are from NLL-based scoring, ratio-based scoring can be an obvious choice where time is an issue.

\bibliography{anthology,custom}

\appendix
\section{Appendix}\label{sec:appendix}
\subsection{Hyper-parameters} \label{sec:appendix_hyperparameter}
\begin{table}[!ht]
    \centering
    \small
    \begin{tabular}{l|c}\toprule
    \multicolumn{2}{c}{\textbf{Training Parameters}}\\ \midrule
    Criterion & Label Smoothing Cross Entropy \\
    Label Smoothing & 0.1 \\
    Optimizer & Adam \\
    Learning Rate (LR) & 2e-3 \\
    LR Scheduler & Inverse Sqrt \\
    Max Tokens & 40000 \\
    Warm-up Updates & 10000 \\
    Max Update & 100000 \\
    Clip Norm & 10.0 \\
    Update Frequency & 8 \\ \bottomrule
    \end{tabular}
    \caption{Hyper-parameters during training our ASR and ST models.}
    \label{tab:hyp_train}
\end{table}
\begin{table}[!ht]
    \centering
    \small
    \begin{tabular}{l|c}\toprule
    \multicolumn{2}{c}{\textbf{Inference Parameters}}\\ \midrule
    Criterion & Label Smoothing Cross Entropy \\
    Beam & 5 \\
    Max Tokens & 10000 \\
    Scoring (ST) & sacrebleu \\
    Scoring (ASR) & WER \\
    \hspace{1em} - WER Tokenizer & 13a \\
    \hspace{1em} - WER Lowercase & True \\
    \hspace{1em} - WER remove Punct & True \\
    \bottomrule
    \end{tabular}
    \caption{Hyper-parameters during inference of our ASR and ST models.}
    \label{tab:hyp_test}
\end{table}
\begin{table}[!ht]
    \centering
    \begin{tabular}{l|c}\toprule
    \multicolumn{2}{c}{\textbf{Model Architecture}}\\ \midrule
    Vocab Size (ASR) & 5000 \\
    Vocab Size (ST) & 8000 \\
    Vocab Type & Unigram \\
    Conv1d Kernel Size & 5,5 \\
    Conv1d Channels & 1024 \\
    Con21d Channels & 256 \\
    Encoder Embedding Dim & 256 \\
    Encoder FFN Embedding Dim & 256 * 8 \\
    Encoder Attention Head & 4 \\
    Encoder Layer & 12 \\
    Decoder Embedding Dim & 256 \\
    Decoder FFN Embedding Dim & 256 * 8 \\
    Decoder Attention Head & 4 \\
    Decoder Layer & 6 \\
    Dropout & 0.1 \\
    Activation Function & RELU \\
    \bottomrule
    \end{tabular}
    \caption{Hyper-parameters of our ASR and ST model architecture.}
    \label{tab:hup_model}
\end{table}

\newpage
\subsection{Bilingual Systems Full Results} \label{sec:appendix_bilingual}
\begin{table*}[!t]\centering
\footnotesize
\begin{tabular}{lrr|lrrr}\toprule
\textbf{ST System (Only SM)} &\textbf{\# of pairs} &\textbf{BLEU} &\textbf{ST System (MuST-C + SM)} &\textbf{\# of pairs} &\textbf{BLEU} &\textbf{Overlap} \\\midrule
 & & &\textbf{~\cite{di-gangi-etal-2019-must}} &269256 &22.29 & \\
 & & &\textbf{(i) Baseline (MuST-C)} &269256 &31.51 & \\
\textbf{Baseline (unfiltered SM)} &1384112 &4.87 &\textbf{(ii) Baseline (MuST-C + unfiltered SM)} &1653367 &28.06 & \\
\midrule
\textbf{20\% NLL} &276823 &6.01 &\textbf{(i) + 20\% NLL} &546078 &\textbf{32.07} &100.00 \\
\textbf{40\% NLL} &553645 &5.41 &\textbf{(i) + 40\% NLL} &822900 &31.86 &66.36 \\
\textbf{60\% NLL} &830467 &7.74 &\textbf{(i) + 60\% NLL} &1099722 &31 &49.66 \\
\textbf{80\% NLL} &1107289 &\textbf{8.42} &\textbf{(i) + 80\% NLL} &1376544 &30.13 &39.67 \\
\midrule
\textbf{0.25 z Text-Text} &251201 &4.64 &\textbf{(i) + 0.25 z Text-Text} &520456 &31.82 &62.14 \\
\textbf{0.5 z Text-Text} &582453 &7.39 &\textbf{(i) + 0.5 z Text-Text} &851708 &31.2 &74.57 \\
\textbf{0.75 z Text-Text} &820294 &7.79 &\textbf{(i) + 0.75 z Text-Text} &1089549 &30.04 &84.66 \\
\textbf{1 z Text-Text} &1017152 &5.54 &\textbf{(i) + 1 z Text-Text} &1286407 &29.52 &91.43 \\
\midrule
\textbf{0.25 z Speech-Speech} &409934 &2.76 &\textbf{(i) + 0.25 z Speech-Speech} &679189 &31.28 &58.38 \\
\textbf{0.5 z Speech-Speech} &723714 &4.63 &\textbf{(i) + 0.5 z Speech-Speech} &992969 &29.95 &66.60 \\
\textbf{0.75 z Speech-Speech} &945962 &5.05 &\textbf{(i) + 0.75 z Speech-Speech} &1215217 &29.6 &73.84 \\
\textbf{1 z Speech-Speech} &1095298 &4.81 &\textbf{(i) + 1 z Speech-Speech} &1364553 &29.13 &79.89 \\
\midrule
\textbf{0.25 z Speech-Text} &458789 &5.77 &\textbf{(i) + 0.25 z Speech-Text} &728044 &31.05 &55.30 \\
\textbf{0.5 z Speech-Text} &860457 &7.3 &\textbf{(i) + 0.5 z Speech-Text} &1129712 &29.64 &67.51 \\
\textbf{0.75 z Speech-Text} &1133782 &7.57 &\textbf{(i) + 0.75 z Speech-Text} &1403037 &29.04 &75.42 \\
\textbf{1 z Speech-Text} &1266324 &6.82 &\textbf{(i) + 1 z Speech-Text} &1535579 &28.89 &82.75 \\
\midrule
\textbf{0.25 z Text-Speech} &311053 &2.13 &\textbf{(i) + 0.25 z Text-Speech} &580308 &31.81 &57.82 \\
\textbf{0.5 z Text-Speech} &601697 &3.88 &\textbf{(i) + 0.5 z Text-Speech} &870952 &30.57 &66.56 \\
\textbf{0.75 z Text-Speech} &849956 &5.23 &\textbf{(i) + 0.75 z Text-Speech} &1119211 &29.91 &74.09 \\
\textbf{1 z Text-Speech} &1042400 &6.21 &\textbf{(i) + 1 z Text-Speech} &1311655 &29.69 &80.55 \\
\midrule
\textbf{} & & &\textbf{(i) + 20\% NLL $\bigcup$ 0.25 z Text-Text} &764533 &31.58 & 100.00\\
\textbf{} & & &\textbf{(i) + 20\% NLL $\bigcap$ 0.25 z Text-Text} &302001 &30.94 & 100.00\\
\bottomrule
\end{tabular}
\caption{Results of filtering approaches for En-Fr direction}
\label{tab:EN-FR}
\end{table*}
\begin{table*}[!t]\centering
\small
\begin{tabular}{lrr|lrrr}\toprule
\textbf{ST System (Only SM)} &\textbf{\# of pairs} &\textbf{BLEU} &\textbf{ST System (MTedX + SM)} &\textbf{\# of pairs} &\textbf{BLEU} &\textbf{Overlap} \\\midrule
\textbf{} & & &\textbf{~\cite{DBLP:journals/corr/abs-2102-01757}} &29946 &8.9 & \\
\textbf{} & & &\textbf{(i) Baseline (MTedX)} &29946 &14.74 & \\
\textbf{Baseline (unfiltered SM)} &1403985 &5.2 &\textbf{(ii) Baseline (MTedX + unfiltered SM)} &1433931 &26.74 & \\
\midrule
\textbf{20\% NLL} &280797 &5.15 &\textbf{(i) + 20\% NLL} &310743 &22.33 &100.00 \\
\textbf{40\% NLL} &561594 &\textbf{6.31} &\textbf{(i) + 40\% NLL} &591540 &25.46 &100.00 \\
\textbf{60\% NLL} &842391 &5.36 &\textbf{(i) + 60\% NLL} &872337 &\textbf{27.66} &100.00 \\
\textbf{80\% NLL} &1123188 &5.68 &\textbf{(i) + 80\% NLL} &1153134 &27.23 &75.65 \\
\midrule
\textbf{0.25 z Text-Text} &475032 &3.8 &\textbf{(i) + 0.25 z Text-Text} &504978 &26.52 &37.18 \\
\textbf{0.5 z Text-Text} &893780 &4.78 &\textbf{(i) + 0.5 z Text-Text} &923726 &26.58 &66.73 \\
\textbf{0.75 z Text-Text} &1159658 &4.11 &\textbf{(i) + 0.75 z Text-Text} &1189604 &26.81 &85.20 \\
\textbf{1 z Text-Text} &1285668 &4.61 &\textbf{(i) + 1 z Text-Text} &1315614 &26.33 &93.89 \\
\midrule
\textbf{0.25 z Speech-Speech} &413584 &4.24 &\textbf{(i) + 0.25 z Speech-Speech} &443530 &26.85 &32.76 \\
\textbf{0.5 z Speech-Speech} &729455 &4.28 &\textbf{(i) + 0.5 z Speech-Speech} &759401 &26.68 &55.00 \\
\textbf{0.75 z Speech-Speech} &952797 &4.51 &\textbf{(i) + 0.75 z Speech-Speech} &982743 &27.06 &70.73 \\
\textbf{1 z Speech-Speech} &1102227 &4.94 &\textbf{(i) + 1 z Speech-Speech} &1132173 &26.62 &81.25 \\
\midrule
\textbf{0.25 z Speech-Text} &308869 &4.92 &\textbf{(i) + 0.25 z Speech-Text} &338815 &25.21 &25.28 \\
\textbf{0.5 z Speech-Text} &600976 &4.36 &\textbf{(i) + 0.5 z Speech-Text} &630922 &26.35 &45.88 \\
\textbf{0.75 z Speech-Text} &849820 &4.56 &\textbf{(i) + 0.75 z Speech-Text} &879766 &26.55 &63.44 \\
\textbf{1 z Speech-Text} &1041531 &4.25 &\textbf{(i) + 1 z Speech-Text} &1071477 &26.7 &76.93 \\
\midrule
\textbf{0.25 z Text-Speech} &458975 &4.37 &\textbf{(i) + 0.25 z Text-Speech} &488921 &26.07 &35.79 \\
\textbf{0.5 z Text-Speech} &861614 &3.4 &\textbf{(i) + 0.5 z Text-Speech} &891560 &26.22 &64.12 \\
\textbf{0.75 z Text-Speech} &1134478 &4.91 &\textbf{(i) + 0.75 z Text-Speech} &1164424 &26.26 &83.33 \\
\textbf{1 z Text-Speech} &1266947 &4.31 &\textbf{(i) + 1 z Text-Speech} &1296893 &26.64 &92.59 \\
\midrule
& & &\textbf{(i) + 60\% NLL $\bigcup$ 0.75 z Speech-Speech} &1250087 &27.18 & 100.00\\
& & &\textbf{(i) + 60\% NLL $\bigcap$ 0.75 z Speech-Speech} &604993 &26.22 & 100.00\\
\bottomrule
\end{tabular}
\caption{Results of filtering approaches for Fr-En direction}
\label{tab:FR-EN}
\end{table*}
\begin{table*}[!t]\centering
\small
\begin{tabular}{lrr|lrrr}\toprule
\textbf{ST System (Only SM)} &\textbf{\# of pairs} &\textbf{BLEU} &\textbf{ST System (MTEdX + SM)} &\textbf{\# of pairs} &\textbf{BLEU} &\textbf{Overlap} \\\midrule
\textbf{} & & &\textbf{~\cite{DBLP:journals/corr/abs-2102-01757}} &20821 &12.2 & \\
\textbf{} & & &\textbf{(i) Baseline (MTedX)} &20821 &21.05 & \\
\textbf{Baseline (unfiltered SM)} &1074027 &0.57 &\textbf{(ii) Baseline (MTedX + unfiltered SM)} &1094848 &26.28 & \\
\midrule
\textbf{20\% NLL} &214806 &2.54 &\textbf{(i) + 20\% NLL} &235627 &24.44 &49.43 \\
\textbf{40\% NLL} &429611 &2.76 &\textbf{(i) + 40\% NLL} &450432 &24.57 &48.60 \\
\textbf{60\% NLL} &644416 &4.07 &\textbf{(i) + 60\% NLL} &665237 &24.45 &49.72 \\
\textbf{80\% NLL} &859221 &4.24 &\textbf{(i) + 80\% NLL} &880042 &26.16 &51.21 \\
\midrule
\textbf{0.25 z Text-Text} &294944 &3.9 &\textbf{(i) + 0.25 z Text-Text} &315765 &25.62 &54.43 \\
\textbf{0.5 z Text-Text} &559285 &\textbf{5.13} &\textbf{(i) + 0.5 z Text-Text} &580106 &26.77 &100.00 \\
\textbf{0.75 z Text-Text} &762804 &4.55 &\textbf{(i) + 0.75 z Text-Text} &783625 &26.22 &100.00 \\
\textbf{1 z Text-Text} &904329 &4.2 &\textbf{(i) + 1 z Text-Text} &925150 &26.13 &100.00 \\
\midrule
\textbf{0.25 z Speech-Speech} &370267 &3.34 &\textbf{(i) + 0.25 z Speech-Speech} &391088 &26.3 &39.95 \\
\textbf{0.5 z Speech-Speech} &584872 &4.09 &\textbf{(i) + 0.5 z Speech-Speech} &605693 &26.67 &60.64 \\
\textbf{0.75 z Speech-Speech} &739279 &4.39 &\textbf{(i) + 0.75 z Speech-Speech} &760100 &26.57 &74.69 \\
\textbf{1 z Speech-Speech} &847397 &4.51 &\textbf{(i) + 1 z Speech-Speech} &868218 &26.35 &83.81 \\
\midrule
\textbf{0.25 z Speech-Text} &295074 &3.68 &\textbf{(i) + 0.25 z Speech-Text} &315895 &24.74 &36.25 \\
\textbf{0.5 z Speech-Text} &570658 &5.02 &\textbf{(i) + 0.5 z Speech-Text} &591479 &26.13 &64.95 \\
\textbf{0.75 z Speech-Text} &788590 &4.95 &\textbf{(i) + 0.75 z Speech-Text} &809411 &26.47 &84.40 \\
\textbf{1 z Speech-Text} &924162 &4.56 &\textbf{(i) + 1 z Speech-Text} &944983 &26.64 &93.82 \\
\midrule
\textbf{0.25 z Text-Speech} &288588 &2.32 &\textbf{(i) + 0.25 z Text-Speech} &309409 &24.98 &34.47 \\
\textbf{0.5 z Text-Speech} &549748 &3.79 &\textbf{(i) + 0.5 z Text-Speech} &570569 &25.55 &60.88 \\
\textbf{0.75 z Text-Speech} &755145 &3.2 &\textbf{(i) + 0.75 z Text-Speech} &775966 &24.96 &79.37 \\
\textbf{1 z Text-Speech} &897009 &2.81 &\textbf{(i) + 1 z Text-Speech} &917830 &25.65 &89.99 \\
\midrule
& & &\textbf{(i) + 0.5 z Text-Text $\bigcup$ 0.5 z Speech-Speech} &834039 &\textbf{27.07} & 100.00 \\
& & &\textbf{(i) + 0.5 z Text-Text $\bigcap$ 0.5 z Speech-Speech} &351760 &26.46 & 100.00 \\
\bottomrule
\end{tabular}
\caption{Results of filtering approaches for Es-Pt direction}
\label{tab:ES-PT}
\end{table*}
\begin{table*}[!t]\centering
\small
\begin{tabular}{lrr|lrrr}\toprule
\textbf{ST System (Only SM)} &\textbf{\# of pairs} &\textbf{BLEU} &\textbf{ST System (MTedX + SM)} &\textbf{\# of pairs} &\textbf{BLEU} &\textbf{Overlap} \\\midrule
\textbf{} & & &\textbf{~\cite{DBLP:journals/corr/abs-2102-01757}} &11353 &8.7 & \\
\textbf{} & & &\textbf{(i) Baseline (MTedX)} &11353 &12.62 & \\
\textbf{Baseline (SM)} &1107283 &3.23 &\textbf{(ii) Baseline (MTedX + unfiltered SM)} &1118635 &22.64 & \\
\midrule
\textbf{20\% NLL} &221457 &3.24 &\textbf{(i) + 20\% NLL} &232810 &20.01 &100.00 \\
\textbf{40\% NLL} &442913 &4.11 &\textbf{(i) + 40\% NLL} &454266 &20.07 &100.00 \\
\textbf{60\% NLL} &664369 &\textbf{4.38} &\textbf{(i) + 60\% NLL} &675722 &21.02 &100.00 \\
\textbf{80\% NLL} &885825 &4.02 &\textbf{(i) + 80\% NLL} &897178 &\textbf{23.7} &100.00 \\
\midrule
\textbf{0.25 z Text-Text} &314739 &1.35 &\textbf{(i) + 0.25 z Text-Text} &326092 &21.66 &30.41 \\
\textbf{0.5 z Text-Text} &603732 &2.98 &\textbf{(i) + 0.5 z Text-Text} &615085 &22.88 &57.11 \\
\textbf{0.75 z Text-Text} &818558 &2.97 &\textbf{(i) + 0.75 z Text-Text} &829911 &22.91 &76.83 \\
\textbf{1 z Text-Text} &950625 &3.53 &\textbf{(i) + 1 z Text-Text} &961978 &23.07 &88.98 \\
\midrule
\textbf{0.25 z Speech-Speech} &363865 &2.2 &\textbf{(i) + 0.25 z Speech-Speech} &375218 &22.51 &35.40 \\
\textbf{0.5 z Speech-Speech} &596387 &2.97 &\textbf{(i) + 0.5 z Speech-Speech} &607740 &22.08 &56.79 \\
\textbf{0.75 z Speech-Speech} &756379 &3.22 &\textbf{(i) + 0.75 z Speech-Speech} &767732 &23.19 &71.46 \\
\textbf{1 z Speech-Speech} &865675 &3.34 &\textbf{(i) + 1 z Speech-Speech} &877028 &22.46 &81.42 \\
\midrule
\textbf{0.25 z Speech-Text} &290424 &2.24 &\textbf{(i) + 0.25 z Speech-Text} &301777 &20.97 &28.18 \\
\textbf{0.5 z Speech-Text} &552538 &2.51 &\textbf{(i) + 0.5 z Speech-Text} &563891 &22.7 &52.39 \\
\textbf{0.75 z Speech-Text} &760190 &2.78 &\textbf{(i) + 0.75 z Speech-Text} &771543 &22.67 &71.52 \\
\textbf{1 z Speech-Text} &901333 &3.88 &\textbf{(i) + 1 z Speech-Text} &912686 &21.88 &84.46 \\
\midrule
\textbf{0.25 z Text-Speech} &295699 &1.27 &\textbf{(i) + 0.25 z Text-Speech} &307052 &20.07 &28.66 \\
\textbf{0.5 z Text-Speech} &571514 &2.77 &\textbf{(i) + 0.5 z Text-Speech} &582867 &22.69 &54.09 \\
\textbf{0.75 z Text-Speech} &790076 &3.29 &\textbf{(i) + 0.75 z Text-Speech} &801429 &23.32 &74.20 \\
\textbf{1 z Text-Speech} &927495 &2.4 &\textbf{(i) + 1 z Text-Speech} &938848 &22.96 &86.83 \\
\midrule
& & &\textbf{(i) + 80\% NLL $\bigcup$ 0.75 z Speech-Speech} &1043876 &22.6 & 100.00\\
& & &\textbf{(i) + 80\% NLL $\bigcap$ 0.75 z Speech-Speech} &621034 &22.94 & 100.00\\
\bottomrule
\end{tabular}
\caption{Results of filtering approaches for Pt-Es direction}
\label{tab:PT-ES}
\end{table*}

\end{document}